\documentclass{article}

\usepackage[preprint]{neurips_2026}

\usepackage[utf8]{inputenc}
\usepackage[T1]{fontenc}
\usepackage{hyperref}
\usepackage{url}
\usepackage{booktabs}
\usepackage{amsfonts}
\usepackage{amsmath}
\usepackage{nicefrac}
\usepackage{microtype}
\usepackage{xcolor}
\usepackage{graphicx}
\usepackage{algorithm}
\usepackage{algpseudocode}

\graphicspath{{figures/}}

\title{Draw This First}

\author{%
  Dazhi Zhong \\
  Krea.ai \\
  Wand Technologies \\
  \texttt{elea@krea.ai} \\
  \And
  Rowan Bradbury \\
  Bradbury Group \\
  Wand Technologies \\
  \texttt{rowan@bradburygroup.org} \\
  \And
  Grant Davis \\
  Krea.ai \\
  Wand Technologies \\
  \texttt{grant@krea.ai} \\
}
\begin{document}

\maketitle

\begin{abstract}
We invert the typical formulation of sketch generation: instead of drawing strokes in order, we predict a 2D field that defines the order in which strokes are drawn. We use a pretrained latent flow-matching transformer to supply the image prior to predict an intermediate representation, while training the VAE's decoder to predict the order field, stroke mask, and stroke segmentation. We vectorize the predicted segmentation into polylines and sort them by the field, producing an ordered vector sketch. Our model can predict an ordered vector sketch from a text description or derender an image into ordered vectors; for either, it follows text instructions specifying the order of drawing.
\end{abstract}

\section{Introduction}

Sketch generation is a diverse field featuring a variety of techniques and output representations. Points or latents can be predicted autoregressively, commonly through a transformer or RNN \citep{strokenuwa, vqsgen, cose, sketchrnn}. Ordered or unordered coordinates and curves can be predicted through diffusion \citep{sketchknitter,strokefusion,swiftsketch, chirodiff} or optimization \citep{clipasso,vectorfusion,svgdreamer}. Video diffusion generates rasterized process frames \citep{videosketcher}.

We propose a model that generates ordered vector strokes, reuses a pretrained image prior, and allows the requested order to be specified in text. We utilize a commissioned dataset of 47{,}318 handmade artist drawings whose quality far exceeds the doodle corpora that existing sequential sketch work is built upon \citep{quickdraw,doodlergan}. We construct an image-native intermediate representation through an order-as-color codec to enable training the DiT generative model within its trained image latent space. To bridge the gap between representation and vector, we train a VAE decoder to read the intermediate, and emit a draw order field, foreground mask, and stroke-instance segmentation embeddings. We apply smoothing and RDP simplification on the predicted field to yield the final vector stroke. To allow the trained model to follow instructions specifying arbitrary order, we train with a constrained permuted order, with the captions programmatically generated to match the permuted order. We utilize bounding box based annotations to classify strokes to textual captions. We train on both an image-derendering objective and a text-conditioned generation objective over permuted order. We evaluate the model on field, segmentation, geometry, and order metrics, and zero-shot text-to-sketch recognition.

We show that this model is capable of faithfully derendering sketches, following order instructions closely, and retaining the strong text-to-sketch generation capabilities of the prior.

\section{Related Work}
\label{sec:related}

Prior systems differ most usefully in their output contract. Autoregression emits points, SVG operations, or learned stroke tokens in sequence \citep{sketchrnn,iconshop,strokenuwa,vqsgen,cose}; diffusion operates over coordinate sequences or per-stroke latents \citep{sketchknitter,swiftsketch,chirodiff,strokefusion}; per-image optimization fits B\'ezier curves through differentiable rasterization \citep{clipasso,diffvg} against pretrained image priors \citep{vectorfusion,svgdreamer}; multimodal LLMs map images or text to SVG code or ordered strokes \citep{inksight,starvector,omnisvg,sketchagent,llm4svg}, and a concurrent agent draws one semantic part at a time via multi-turn RL over part-annotated sketches \citep{sketchonepart}; video models produce raster process frames rather than vectors \citep{videosketcher}; part-sequential generators \citep{doodlergan,doodleformer} and implicit-time representations \citep{sketchinr,sketchode} round out the space. Across all of these, the drawing process is either the model's own generation axis or not represented; none expose order as a free-form language-conditioned variable. Offline-to-online handwriting conversion supplies order-recovery metrics -- RMSE, SNR, DTW against recorded trajectories \citep{wor,setsort} -- which inform our evaluation. Our backbone is Qwen-Image-Edit-2509 \citep{qwenimage}: a VAE compressing pixels to a latent grid and a diffusion transformer trained by latent flow matching \citep{flowmatching}, adapted with LoRA \citep{lora}.

\section{Method}

\subsection{Sketch dataset, hierarchical annotation, order captions, and permutation}
\label{sec:aug}

We gather a corpus of 47{,}318 drawings, commissioned through 50 distinct artists on Upwork. Each drawing is stored as the raw Apple Pencil input stream plus a simplified per-point form $(x, y)$ forming a sequence of polylines, in this work we only use the simplified form. Drawings average 77.5 strokes and 6{,}864 points (median 57 and 5{,}548; p99 387 and 26{,}028).

Each drawing is annotated with a two-level bounding-box hierarchy, structured as region then subject (Figure~\ref{fig:annotation}). The drawing order from the dataset is commonly non continuous and contain many revisits, where additional detail is added to already drawn regions. We synthesize order-descriptive captions as a sequence of visits through named regions, specifying locale upon revisit, for example: ``box1 left, then box2, then box1 right, revisited.'' We then teach the model to generate conditioned on order by permuting the target's stroke order and generate the matching caption, forming a order supervised pair. When permuting, \textbf{within-region order is preserved} while varying between-region order, so the \textbf{permutation keeps the natural order of drawing from rough outline to details}. We augment the image condition given to the model as reference, while preserving pixel perfect stroke mapping (Figure~\ref{fig:dit-architecture}a).

\subsection{Intermediate image representation with order-as-color encoding}
\label{sec:hsv}

Let a sketch $S = \{s_i\}_{i=1}^{N}$ be a sequence of strokes, each a sequence of points $s_i = \{p_{i,j}\}_{j=1}^{M_i}$, $p_{i,j} \in \mathbb{R}^2$. Write $\ell(p_{i,j})$ for arc length accumulated from the first point of $s_1$ to $p_{i,j}$ along the drawing, $L$ for the total, and $\ell_i(p_{i,j})$ for arc length within $s_i$ alone. The global arc $a = \ell/L \in [0,1]$ is a single monotone scalar which represents the drawing's progress. The within-stroke arc is $u = \ell_i / \ell_i(p_{i,M_i})$. The sketch $X = \mathrm{enc}(S)$ is encoded at each inked pixel,
\begin{equation}
H = a \cdot \tfrac{342}{360}, \qquad
S_{\text{sat}} = 1, \qquad
V = 1 - \tfrac{1}{2} u ,
\end{equation}
with $\mathrm{HSV}(0,0,1)$ (white) on background (Figure~\ref{fig:codec}). This rasterized sketch is referred as the codec image. This encoding produces single-pixel polylines, and recorded brush width is not encoded.

\subsection{Order-native decoder}
\label{sec:decoder}

\begin{figure}[t]
\centering
\includegraphics[width=\linewidth]{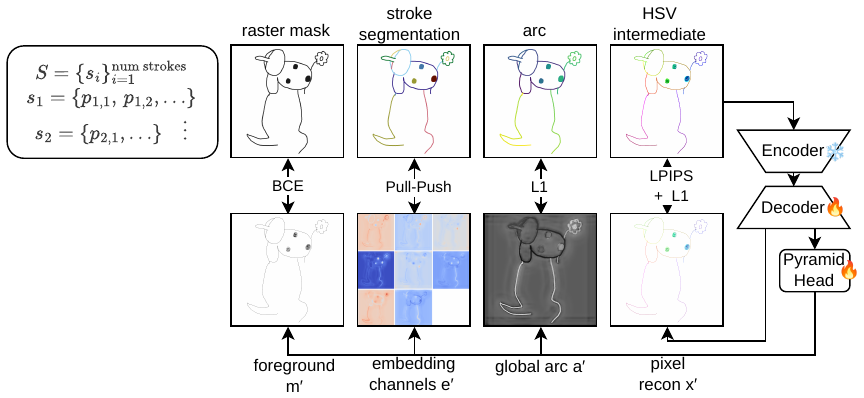}
\caption{Decoder architecture and objective. The source strokes derive four targets: ink raster, stroke segmentation, arc field, and the intermediate HSV image. The frozen encoder and trained decoder trunk with pyramid head emit foreground, instance embedding (all eight raw channels shown), arc, and pixel outputs; each loss bridges an output to its target.}
\label{fig:decoder}
\end{figure}

We finetune Qwen-Image's VAE's decoder for better stroke field recovery, the encoder $E$ stays frozen (Figure~\ref{fig:decoder}). The supervised target is to reconstruct the global arc field, mask, and stroke segmentation given the latent of the encoded intermediate representation image. The decoder trunk is retrained jointly with new heads: a CNN head is added to the trunk's penultimate features, structured as a pyramid at full, $4\times$, and $16\times$ resolution; a final convolution layer reads the channel-concatenated features and emits 10 channels: global arc $a'$, foreground logit $m'$, and an eight-dimensional instance embedding $e'$. The pixel branch is architecturally unchanged, while training jointly with the new head with a stroke-weighted L1 loss and perceptual loss; the full objective and weights are in Appendix~\ref{app:decoder}. We train this decoder for 30{,}000 steps.

\subsection{Generative model and stroke recovery}
\label{sec:dit}

The generator is Qwen-Image-Edit-2509 \citep{qwenimage} adapted with a rank-64 LoRA \citep{lora} ($\alpha = 64$), trained 50{,}000 steps on 32 H100-80GB GPUs (2,848 GPU hours) with the standard latent flow-matching objective on the latent of the intermediate representation, $z = E(\mathrm{enc}(S))$, where the condition, text or image, is drawn from the permuted and augmented distribution of Section~\ref{sec:aug} (Figure~\ref{fig:dit-architecture}). Optimization details are in Appendix~\ref{app:decoder}. Both trainings are implemented in bbml \citep{bbml}.

\begin{figure}[t]
\centering
\includegraphics[width=\linewidth]{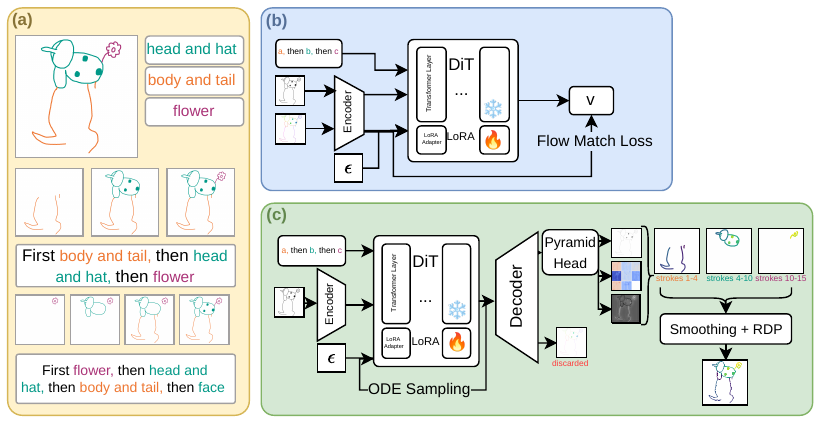}
\caption{(a) Order permutation with local order preservation from Section~\ref{sec:aug}. (b) The DiT trains in the latent space of the codec image. (c) At inference the predicted latent feeds the order-native decoder; the pixel output is discarded.}
\label{fig:dit-architecture}
\end{figure}

We convert the decoder emitted fields to ordered polyline paths without a learned tracer. HDBSCAN clusters foreground embeddings $e'$ to segmentations. Clusters are globally sorted by mean predicted arc $a'$. Within each cluster, a nearest-neighbour walks foreground $m'$ pixels, splitting into subsegments then merging into full vector strokes, with vertices aligned to the $1024\times1024$ grid. These strokes are then smoothed, then simplified with Ramer-Douglas-Peucker with epsilon 0.5\,px. The order inside and between strokes are determined by global arc $a'$ (Algorithm~\ref{alg:decode} in the appendix). This polyline output is the final vectorized result.

\section{Results}
\label{sec:results}

We evaluate 100 drawings from each of our dataset's training holdout, Creative Birds, Creative Creatures \citep{doodlergan}, and FS-COCO \citep{fscoco}; the latter three never enter DiT training. The primary order metric is Kendall $\tau$ between predicted and recorded stroke order after Hungarian geometric matching of strokes; since several orders are plausible per drawing, $\tau$ against the single recorded order is conservative. As controls, randomly permuting predicted order drives matched $\tau$ to $0.02$--$0.04$ and full reversal to $-1.0$ on all sets, validating the metric.
Where noted we also report DTW over stroke trajectories, resampled uniformly to 128 points, and accumulated cost is divided by path length.

\subsection{The decoder ceiling}

Decoding latents of encoded ground-truth codec images isolates losses from the encoder, decoder, and vectorization stack. The order-native head cuts arc $L_1$ by $9$--$19\times$ relative to reading hue from the frozen decoder, with arc Spearman $\geq 0.994$ and mask IoU $\geq 0.97$ on all five datasets (full tables in Appendix~\ref{app:ceiling}). Carried through segmentation and polyline fitting, the deployment ceiling reaches matched Kendall $\tau$ of $0.91$--$0.94$ on the four multi-stroke datasets ($0.56$ on QuickDraw, whose sparse short strokes fragment).

\subsection{Order is written by language}
\label{sec:instructions}

\begin{figure}[t]
\centering
\includegraphics[width=0.9\linewidth]{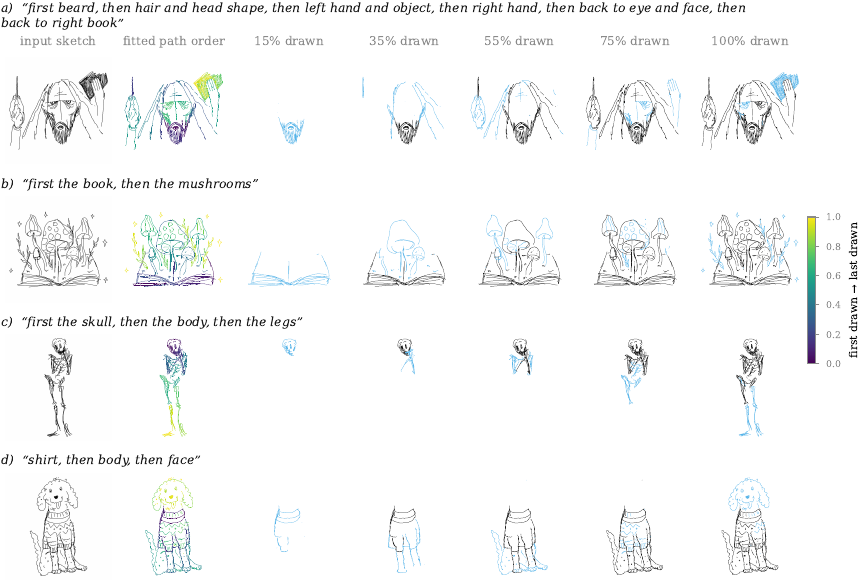}
\caption{Draw order under text intervention: four held-out sketches, each derendered under a hand-written order instruction. Fitted path order (viridis, first to last) and replay at $15/35/55/75/100\%$ -- black: already drawn, blue: added since the previous frame. Guidance 10.}
\label{fig:orderprompt}
\end{figure}

Table~\ref{tab:intervention} measures the instruction's effect on predicted order. Geometry never moves under any intervention while order swings across its full range: deleting the caption drops in-domain order to the level of geometry-only baselines (Appendix~\ref{app:uncond}), stating the recorded order lifts out-of-domain adherence, and reversing the same instruction inverts it -- the caption, not the condition image, is the order channel, and the model follows a stated order even when it contradicts the recording.
Figure~\ref{fig:orderprompt} shows how instructions steer the produced order on held-out sketches.

\begin{table}[t]
\centering
\caption{Caption interventions under fixed inputs and seed ($n=100$ per row). Top panel: in-domain holdout, caption deleted. Bottom panel: Creative Birds+Creatures -- captionless, recorded part order stated, or reversed. Order metrics are measured against the recorded order; geometry columns show the intervention leaves the drawing itself unchanged.}
\label{tab:intervention}
\small
\begin{tabular}{llrrr}
\toprule
panel & caption & Kendall $\tau$ & stroke match & mask IoU \\
\midrule
holdout (in-domain)
 & no caption& 0.096  & 0.896 & 0.775 \\
 & with caption& 0.449  & 0.895 & 0.774 \\
\midrule
Birds+Creatures (out-of-domain)
 & no caption          & 0.139  & 0.941 & 0.599 \\
 & recorded order      & 0.373  & 0.934 & 0.601 \\
 & reversed order      & $-0.267$ & 0.934 & 0.601 \\
\bottomrule
\end{tabular}
\end{table}

\subsection{The image prior survives the finetune}

The backbone's open-vocabulary drawing ability is retained allowing generation of any ordered sketch with the original model's world knowledge. Prompted with 50 QuickDraw categories it never saw as vectors, the model's final rendered SVGs are recognized by CLIP ViT-B/32 at Top-1 $0.70$ (guidance 5.0) and $0.75$ (7.5), Top-5 $0.84$ (Table~\ref{tab:clip-cfg}). We showcase a representative sample in Figure~\ref{fig:results}, last row.

\begin{table}[t]
\centering
\caption{QuickDraw text-to-vector recognition, CLIP ViT-B/32 over 50 categories $\times$ two seeds.}
\label{tab:clip-cfg}
\small
\begin{tabular}{rrrrr}
\toprule
CFG & Top-1 $\uparrow$ & Top-5 $\uparrow$ & target rank $\downarrow$ & target probability $\uparrow$ \\
\midrule
5.0 & 0.70 & 0.78 & 6.23 & 0.640 \\
7.5 & \textbf{0.75} & \textbf{0.84} & \textbf{4.62} & \textbf{0.692} \\
\bottomrule
\end{tabular}
\end{table}

\section{How precise is order control?}
\label{sec:granularity}

Order instructions name units such as regions, subjects, parts,  while leaving order inside them unspecified. We score against the instruction's specifications by assigning predicted order to ground-truth strokes. We report Kendall $\tau$ between predicted order and instructed order, per unit over the strokes the instruction constrains. Residual order \emph{inside} units is reported separately as within-unit $\tau$.

\paragraph{Adherence degrades with instruction granularity.}
On the holdout test set we issue permuted order instructions at the three trained caption shapes, instructing top level region order, subject order, and revisit-instructing passes (Table~\ref{tab:precedence}). Region-level instructions are followed reliably, adherence falls to $0.46$ at part level and $0.44$ at pass level.

\paragraph{An external part-annotated test.}
ControlSketch-Part \citep{sketchonepart} pairs each synthetic 32-stroke sketch with 2 to 5 named parts assigned to each stroke. We derender 100 test sketches from categories disjoint from both its training split and ours. Given order instructions, $\tau$ reaches $0.78$ as stated and $0.73$ \emph{reversed}, from an uninstructed baseline of $0.36$; mean per-sample Spearman between instructed and produced part order is $0.90$ and $0.87$.

\begin{table}[t]
\centering
\caption{Stroke-level Kendall $\tau$ against each cell's own instruction, over instruction-constrained stroke pairs ($n=100$). Parenthesized: mean instructed units per drawing. Within-unit $\tau$: residual stroke order inside instructed units.}
\label{tab:precedence}
\small
\begin{tabular}{llrr}
\toprule
panel & instruction & stroke $\tau$ & within-unit $\tau$ \\
\midrule
holdout            & none (ordinary caption)  & 0.149 & 0.382 \\
                   & permuted regions (2.8)   & 0.838 & 0.282 \\
                   & permuted parts (7.1)     & 0.461 & 0.177 \\
                   & permuted passes (14.9)   & 0.444 & 0.134 \\
\midrule
ControlSketch-Part & none (ordinary caption)  & 0.363 & 0.088 \\
                   & stated part order        & 0.778 & 0.058 \\
                   & reversed part order      & 0.734 & 0.068 \\
\bottomrule
\end{tabular}
\end{table}

\section{Limitations}
\label{sec:limits}

Training permutes stroke part order and rewrites the caption to match, deliberately decorrelating order from geometry; the direct cost is that, given no instruction, the model has no verifiable human-like global order prior. Due to our training captioning schema, the order of objects in the caption text influences the draw order, and generalization to instructed order conflicting to order of appearance is reduced. An explicit order clause survives an adjacent description at a measured cost ($\tau$ $0.78 \to 0.73$ when the stated order matches the description's mention order, $0.73 \to 0.47$ when it opposes it), and a single trailing constraint is lost entirely (named region drawn first $0.315$ of the time appended to the full caption, vs.\ $0.326$ uninstructed and $0.517$ for the same sentence alone). Below named units there is no control at all: within-unit order agreement is near zero under every instruction (Table~\ref{tab:precedence}), and inverting a region's trained pass convention succeeds well under half the time ($0.400$, against an uninstructed $0.267$). The amount of control is constrained by the unit-size which the permutation training used.

The decoder and vectorization causes fragmentation. Recovered paths run $1.7$--$2.5\times$ the true stroke count, so strokes are recovered in pieces. Our objective drops recorded time, pressure, tilt, and brush width; predicted paths are 1\,px unweighted centerlines parameterized by cumulative arc length rather than time. Finally, our out-of-domain sets are doodle-flavored; adherence on complex real line drawings is so far only qualitative.

\section{Conclusion}
We present a system that generates ordered vector sketches by rendering draw order into color, generating that image with a pretrained diffusion transformer, and reading order back out with an order-native decoder into replayable polyline paths. Trained on 47{,}318 commissioned artist drawings, the model derenders sketches faithfully and retains the prior's open-vocabulary drawing ability.  On held-out data the model recovers substantial final-vector order. Text is the order channel: captions carry most in-domain order, recorded-order instructions improve it, and reversed instructions invert it, all without changing geometry.

\medskip
{
\small
\bibliographystyle{plainnat}
\bibliography{arxiv}

@inproceedings{chirodiff,
  title     = {ChiroDiff: Modelling Chirographic Data with Diffusion Models},
  author    = {Das, Ayan and Yang, Yongxin and Hospedales, Timothy and Xiang, Tao and Song, Yi-Zhe},
  booktitle = {International Conference on Learning Representations (ICLR)},
  year      = {2023},
  note      = {arXiv:2304.03785},
}

@inproceedings{vectorfusion,
  title     = {VectorFusion: Text-to-SVG by Abstracting Pixel-Based Diffusion Models},
  author    = {Jain, Ajay and Xie, Amber and Abbeel, Pieter},
  booktitle = {Conference on Computer Vision and Pattern Recognition (CVPR)},
  year      = {2023},
  note      = {arXiv:2211.11319},
}

@inproceedings{svgdreamer,
  title     = {SVGDreamer: Text Guided SVG Generation with Diffusion Model},
  author    = {Xing, Ximing and Zhou, Haitao and Wang, Chuang and Zhang, Jing and Xu, Dong and Yu, Qian},
  booktitle = {Conference on Computer Vision and Pattern Recognition (CVPR)},
  year      = {2024},
  note      = {arXiv:2312.16476},
}

@inproceedings{sketchrnn,
  title  = {A Neural Representation of Sketch Drawings},
  author = {Ha, David and Eck, Douglas},
  booktitle = {International Conference on Learning Representations (ICLR)},
  year   = {2018},
  note   = {arXiv:1704.03477},
}

@article{iconshop,
  title   = {IconShop: Text-Guided Vector Icon Synthesis with Autoregressive Transformers},
  author  = {Wu, Ronghuan and Su, Wanchao and Ma, Kede and Liao, Jing},
  journal = {ACM Transactions on Graphics (SIGGRAPH Asia)},
  year    = {2023},
  note    = {arXiv:2304.14400},
}

@inproceedings{strokenuwa,
  title     = {StrokeNUWA: Tokenizing Strokes for Vector Graphic Synthesis},
  author    = {Tang, Zecheng and others},
  booktitle = {International Conference on Machine Learning (ICML)},
  year      = {2024},
  note      = {arXiv:2401.17093},
}

@inproceedings{vqsgen,
  title     = {VQ-SGen: A Vector Quantized Stroke Representation for Creative Sketch Generation},
  author    = {Wang, Jiawei and others},
  booktitle = {International Conference on Computer Vision (ICCV)},
  year      = {2025},
  note      = {arXiv:2411.16446},
}

@inproceedings{cose,
  title     = {CoSE: Compositional Stroke Embeddings},
  author    = {Aksan, Emre and Deselaers, Thomas and Tagliasacchi, Andrea and Hilliges, Otmar},
  booktitle = {Advances in Neural Information Processing Systems (NeurIPS)},
  year      = {2020},
  note      = {arXiv:2006.09930},
}

@article{inksight,
  title   = {InkSight: Offline-to-Online Handwriting Conversion by Learning to Read and Write},
  author  = {Mitrevski, Blagoj and others},
  journal = {Transactions on Machine Learning Research (TMLR)},
  year    = {2024},
  note    = {arXiv:2402.05804},
}

@inproceedings{starvector,
  title     = {StarVector: Generating Scalable Vector Graphics Code from Images and Text},
  author    = {Rodriguez, Juan A. and others},
  booktitle = {Conference on Computer Vision and Pattern Recognition (CVPR)},
  year      = {2025},
  note      = {arXiv:2312.11556},
}

@inproceedings{omnisvg,
  title     = {OmniSVG: A Unified Scalable Vector Graphics Generation Model},
  author    = {Yang, Yiying and others},
  booktitle = {Advances in Neural Information Processing Systems (NeurIPS)},
  year      = {2025},
  note      = {arXiv:2504.06263},
}

@inproceedings{sketchagent,
  title     = {SketchAgent: Language-Driven Sequential Sketch Generation},
  author    = {Vinker, Yael and others},
  booktitle = {Conference on Computer Vision and Pattern Recognition (CVPR)},
  year      = {2025},
  note      = {arXiv:2411.17673},
}

@inproceedings{llm4svg,
  title     = {Empowering LLMs to Understand and Generate Complex Vector Graphics},
  author    = {Xing, Ximing and others},
  booktitle = {Conference on Computer Vision and Pattern Recognition (CVPR)},
  year      = {2025},
  note      = {arXiv:2412.11102},
}

@inproceedings{sketchknitter,
  title     = {SketchKnitter: Vectorized Sketch Generation with Diffusion Models},
  author    = {Wang, Qiang and others},
  booktitle = {International Conference on Learning Representations (ICLR)},
  year      = {2023},
}

@inproceedings{strokefusion,
  title     = {StrokeFusion: Vector Sketch Generation via Joint Stroke-UDF Encoding and Latent Sequence Diffusion},
  author    = {Zhou, Jin and Zhou, Yi and Yang, Hongliang and Xu, Pengfei and Huang, Hui},
  booktitle = {AAAI Conference on Artificial Intelligence},
  year      = {2026},
  note      = {arXiv:2503.23752},
}

@inproceedings{swiftsketch,
  title     = {SwiftSketch: A Diffusion Model for Image-to-Vector Sketch Generation},
  author    = {Arar, Ellie and Frenkel, Yarden and Cohen-Or, Daniel and Shamir, Ariel
               and Vinker, Yael},
  booktitle = {SIGGRAPH},
  year      = {2025},
  note      = {arXiv:2502.08642},
}

@article{clipasso,
  title   = {CLIPasso: Semantically-Aware Object Sketching},
  author  = {Vinker, Yael and others},
  journal = {ACM Transactions on Graphics (SIGGRAPH)},
  volume  = {41},
  number  = {4},
  year    = {2022},
}

@article{diffvg,
  title   = {Differentiable Vector Graphics Rasterization for Editing and Learning},
  author  = {Li, Tzu-Mao and Luk{\'a}{\v{c}}, Michal and Gharbi, Micha{\"e}l and Ragan-Kelley, Jonathan},
  journal = {ACM Transactions on Graphics (SIGGRAPH Asia)},
  volume  = {39},
  number  = {6},
  year    = {2020},
}

@inproceedings{sketchinr,
  title     = {SketchINR: A First Look into Sketches as Implicit Neural Representations},
  author    = {Bandyopadhyay, Hmrishav and Bhunia, Ankan Kumar and Chowdhury, Pinaki Nath
               and Sain, Aneeshan and Xiang, Tao and Song, Yi-Zhe},
  booktitle = {Conference on Computer Vision and Pattern Recognition (CVPR)},
  year      = {2024},
  note      = {arXiv:2403.09344},
}

@inproceedings{sketchode,
  title     = {SketchODE: Learning Neural Sketch Representation in Continuous Time},
  author    = {Das, Ayan and others},
  booktitle = {International Conference on Learning Representations (ICLR)},
  year      = {2022},
}

@article{videosketcher,
  title   = {{VideoSketcher}: Sequential Sketch Generation Using Video Model Priors},
  author  = {Ren, Hui and Alaluf, Yuval and Bar-Tal, Omer and Schwing, Alexander
             and Torralba, Antonio and Vinker, Yael},
  journal = {arXiv preprint arXiv:2602.15819},
  year    = {2026},
}

@inproceedings{doodlergan,
  title     = {Creative Sketch Generation},
  author    = {Ge, Songwei and Goswami, Vedanuj and Zitnick, C. Lawrence and Parikh, Devi},
  booktitle = {International Conference on Learning Representations (ICLR)},
  year      = {2021},
  note      = {arXiv:2011.10039; DoodlerGAN},
}

@article{doodleformer,
  title   = {DoodleFormer: Creative Sketch Drawing with Transformers},
  author  = {Bhunia, Ankan Kumar and Khan, Salman and Cholakkal, Hisham and Anwer, Rao Muhammad
             and Khan, Fahad Shahbaz and Laaksonen, Jorma and Felsberg, Michael},
  journal = {arXiv preprint arXiv:2112.03258},
  year    = {2021},
}

@article{wor,
  title   = {Writing Order Recovery in Complex and Long Static Handwriting},
  author  = {Diaz, Moises and Crispo, Gioele and Parziale, Antonio and Marcelli, Angelo
             and Ferrer, Miguel A.},
  journal = {arXiv preprint arXiv:2406.03194},
  year    = {2024},
}

@inproceedings{setsort,
  title     = {{SET, SORT!} A Novel Sub-Stroke Level Transformers for Offline Handwriting to Online Conversion},
  author    = {Mohamed Moussa, Elmokhtar and Lelore, Thibault and Mouch{\`e}re, Harold},
  booktitle = {International Conference on Document Analysis and Recognition (ICDAR)},
  pages     = {81--97},
  year      = {2023},
  note      = {hal-04182547},
}

@article{qwenimage,
  title   = {Qwen-Image Technical Report},
  author  = {Wu, Chenfei and others},
  journal = {arXiv preprint arXiv:2508.02324},
  year    = {2025},
}

@inproceedings{lora,
  title     = {LoRA: Low-Rank Adaptation of Large Language Models},
  author    = {Hu, Edward J. and Shen, Yelong and Wallis, Phillip and Allen-Zhu, Zeyuan
               and Li, Yuanzhi and Wang, Shean and Wang, Lu and Chen, Weizhu},
  booktitle = {International Conference on Learning Representations (ICLR)},
  year      = {2022},
}

@inproceedings{flowmatching,
  title     = {Flow Matching for Generative Modeling},
  author    = {Lipman, Yaron and Chen, Ricky T. Q. and Ben-Hamu, Heli and Nickel, Maximilian and Le, Matt},
  booktitle = {International Conference on Learning Representations (ICLR)},
  year      = {2023},
}

@misc{quickdraw,
  title  = {The Quick, Draw! Dataset},
  author = {Jongejan, Jonas and Rowley, Henry and Kawashima, Takashi and Kim, Jongmin and Fox-Gieg, Nick},
  year   = {2016},
  note   = {Google Creative Lab},
}

@inproceedings{fscoco,
  title     = {FS-COCO: Towards Understanding of Freehand Sketches of Common Objects in Context},
  author    = {Chowdhury, Pinaki Nath and others},
  booktitle = {European Conference on Computer Vision (ECCV)},
  year      = {2022},
}

@article{sketchonepart,
  title   = {Teaching an Agent to Sketch One Part at a Time},
  author  = {Du, Xiaodan and Xu, Ruize and Yunis, David and Vinker, Yael and Shakhnarovich, Greg},
  journal = {arXiv preprint arXiv:2603.19500},
  year    = {2026},
}

@misc{bbml,
  title        = {bbml: A Better Basic Machine Learning Framework},
  author       = {Bradbury, Rowan and Zhong, Elea},
  year         = {2025},
  howpublished = {\url{https://github.com/Bradbury-Group/bbml}},
}
}

\appendix

\section{Decoder objective and training details}
\label{app:decoder}

Writing $z = E(x)$, $x' = D_p(z)$, and $(a',m',e') = D_{\text{ord}}(z)$,
\begin{equation}
\begin{aligned}
\mathcal{L} ={}&
\underbrace{w_{1} \, \big\| w_{\text{fg}}\!\odot\!(x' - x) \big\|_1}_{\text{foreground-weighted pixel } L_1}
\\
&+ w_{\text{lpips}} \, \mathrm{LPIPS}(x', x)
+ w_{\text{arc}}\Big[
\big\| (a' - a)\!\odot\! m \big\|_1
+ 0.25\,\mathrm{BCE}_{w}(m',m)
+ 0.1\,\mathcal{L}_{\mathrm{pp}}(e',s)
\Big],
\end{aligned}
\end{equation}
where $s$ is the ground-truth stroke-id map, $w_1 = w_{\text{lpips}} = 1$, $w_{\text{arc}} = 8$; stroke pixels and their radius-1 surround receive $50\times$ pixel weight, and the mask BCE uses the same boundary ring with a dynamic positive-class weight. $\mathcal{L}_{\mathrm{pp}}$ is discriminative push--pull: embeddings contract toward their stroke mean outside radius $d_v = 0.5$; stroke means separate by margin $2d_d = 3.0$. Figure~\ref{fig:decoder} shows the architecture.

The DiT LoRA trains with AdamW ($\beta = (0.9, 0.95)$, weight decay $0.01$), learning rate $10^{-4}$ with 100 warmup steps and 10{,}000 steps of decay, global batch size 32, gradient clipping $1.0$, for 50{,}000 steps (Figure~\ref{fig:dit-architecture}).

\begin{algorithm}[h]
\caption{Image-to-vector deployment path}
\label{alg:decode}
\begin{algorithmic}[1]
\Require condition image $c$, caption $q$, generator $G$, decoder $D$
\State $\hat z \gets G(c,q)$
\State $(x',a',m',e') \gets D(\hat z)$
\State $\mathcal{I} \gets \textsc{HDBSCAN}(\{e'(p):m'(p)=1\})$
\State assign embedding-noise pixels to nearest instance centroid
\State $\pi \gets$ instances sorted by mean $a'$
\ForAll{instances $I$ in order $\pi$}
  \State $\mathcal{S}_I \gets \textsc{CappedRadiusWalk}(I)$
  \ForAll{segments $s \in \mathcal{S}_I$}
    \State orient $s$ from lower to higher endpoint arc
  \EndFor
  \State $\mathcal{C}_I \gets \textsc{JoinNearestEndpoints}(\mathcal{S}_I)$
  \State $\mathcal{B}_I \gets \{\textsc{FitPolylineChain}(c):c\in\mathcal{C}_I\}$
\EndFor
\State \Return ordered vectors
$\mathcal{B}_{\pi_1},\ldots,\mathcal{B}_{\pi_{|\mathcal{I}|}}$
\end{algorithmic}
\end{algorithm}

\section{Measuring unconditional generation order}
\label{app:uncond}

\begin{figure}[h]
\centering
\includegraphics[width=\linewidth]{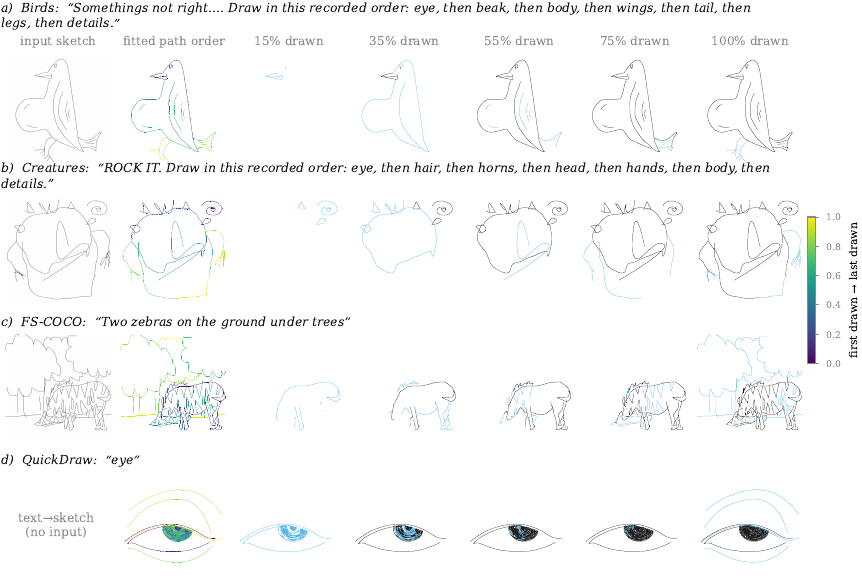}
\caption{Representative 50k outputs, deterministic median selections per domain. (a) Birds and (b) Creatures condition on the input sketch with the recorded part order appended to the caption; (c) FS-COCO uses its plain scene caption (the dataset records no order text); (d) QuickDraw is text-to-sketch, no input image. Columns: input sketch, fitted path order, then cumulative progress frames (black = existing strokes, blue = strokes added since the previous frame).}
\label{fig:results}
\end{figure}

Table~\ref{tab:uncond} reports the full pipeline without order instruction, against two built-in anchors: four deterministic geometry-only orderings of the model's own fitted paths (top-to-bottom, longest-first, centre-out, greedy pen travel; best reported) and a per-drawing retroactive \emph{oracle} over the four. The holdout row deletes the caption, since the holdout's native caption narrates the recorded drawing process (Section~\ref{sec:instructions}); the other sets' native captions carry no order text. Uninstructed, in-domain order sits at the heuristic level ($0.096$ vs.\ $0.105$); on FS-COCO generated order beats the best heuristic by $+0.11$ ($p<10^{-3}$, Wilcoxon); on Birds and Creatures the gap is insignificant. An DTW control agrees: on Birds and Creatures generated order is indistinguishable from paired random order ($0.189$ vs.\ $0.188$, $0.182$ vs.\ $0.180$; lower is better), on FS-COCO it is better ($0.200$ vs.\ $0.227$).
Comparing against the decoder ceiling attributes the pipeline's losses: fragmentation is decode-side (in-domain path-count ratio $2.29$ at ceiling vs.\ $2.27$ end-to-end with the native caption), while order is generation-side -- with the recorded-process caption, end-to-end $\tau$ reaches $0.449$ (Table~\ref{tab:intervention}) against a $0.91$ decode ceiling. Recorded artist order is also simply hard to guess from geometry -- on ground-truth strokes the best heuristic reaches $\tau = 0.137$ and the oracle $0.382$ -- and dense drawings are the hard case even when instructed: under the native caption the simplest holdout quartile (4--26 strokes) reaches $\tau = 0.59$, the busiest (91--379) $0.37$. Figure~\ref{fig:results} shows deterministic median samples per domain.

\begin{table}[t]
\centering
\caption{End-to-end results at 50k steps without order instruction ($n=100$ per dataset). Holdout is generated with the caption deleted; the other sets use their native captions, which carry no order text. Heuristic rows re-order the model's own fitted paths under the identical matching; oracle is the retroactive per-drawing best of the four heuristics. Bold: arc order significantly above the best heuristic (Wilcoxon).}
\label{tab:uncond}
\small
\begin{tabular}{lrrrr}
\toprule
& holdout & Birds & Creatures & FS-COCO \\
\midrule
mask IoU $\uparrow$                & 0.775 & 0.600 & 0.604 & 0.646 \\
stroke match rate $\uparrow$       & 0.896 & 0.922 & 0.955 & 0.795 \\
path-count ratio $\to 1$           & 2.53  & 1.76  & 2.32  & 1.60 \\
Chamfer $\times 10^3$ $\downarrow$ & 0.60  & 0.45  & 0.52  & 0.61 \\
\midrule
Kendall $\tau$, generated arc order $\uparrow$ & 0.096 & 0.256 & 0.213 & \textbf{0.308} \\
Kendall $\tau$, best geometric heuristic       & 0.105 & 0.223 & 0.190 & 0.201 \\
Kendall $\tau$, heuristic oracle               & 0.374 & 0.409 & 0.355 & 0.308 \\
\bottomrule
\end{tabular}
\end{table}

\clearpage
\section{Decoder ceiling tables}
\label{app:ceiling}

\begin{table}[h]
\centering
\caption{Decoder field ceiling from encoded ground-truth latents ($n=100$ each). Frozen reads arc from hue and mask from saturation; Embed is the final order-native decoder. ``Head'' is its calibrated explicit foreground output.}
\label{tab:decoder-field}
\scriptsize
\begin{tabular}{lrrrrrrr}
\toprule
& \multicolumn{2}{c}{arc $L_1$ $\downarrow$}
& \multicolumn{2}{c}{arc Spearman $\uparrow$}
& \multicolumn{2}{c}{saturation mask IoU $\uparrow$}
& head IoU $\uparrow$ \\
dataset & Frozen & Embed & Frozen & Embed & Frozen & Embed & Embed \\
\midrule
holdout   & 0.0607 & 0.0071 & 0.8072 & 0.9937 & 0.8049 & 0.9705 & 0.9706 \\
Birds     & 0.0517 & 0.0027 & 0.8329 & 0.9995 & 0.8188 & 0.9979 & 0.9980 \\
Creatures & 0.0468 & 0.0035 & 0.8670 & 0.9984 & 0.8132 & 0.9955 & 0.9957 \\
FS-COCO   & 0.0605 & 0.0036 & 0.7825 & 0.9983 & 0.8700 & 0.9952 & 0.9952 \\
QuickDraw & 0.0448 & 0.0030 & 0.8675 & 0.9995 & 0.8304 & 0.9986 & 0.9987 \\
\bottomrule
\end{tabular}
\end{table}

\begin{table}[h]
\centering
\caption{Deployment ceiling of the final decoder, including embedding segmentation and polyline fitting ($n=100$).}
\label{tab:decoder-vector}
\small
\begin{tabular}{lrrrrr}
\toprule
& seg.\ ARI & seg.\ IoU & path ratio & match & Kendall $\tau$ \\
\midrule
holdout   & 0.656 & 0.603 & 2.29 & 0.907 & 0.909 \\
Birds     & 0.892 & 0.747 & 1.60 & 0.938 & 0.944 \\
Creatures & 0.840 & 0.720 & 1.93 & 0.955 & 0.941 \\
FS-COCO   & 0.698 & 0.618 & 1.47 & 0.783 & 0.912 \\
QuickDraw & 0.588 & 0.763 & 4.37 & 0.977 & 0.561 \\
\bottomrule
\end{tabular}
\end{table}

\section{Condition funnel details}

\begin{table}[h]
\centering
\caption{Condition funnel: 20 image lanes and 8 caption rungs over 47{,}190 drawings. \emph{pass} is image-alone sufficiency; \emph{+text} additionally credits the paired caption.}
\label{tab:funnel}
\small
\begin{tabular}{lllrr}
\toprule
category & lane & rows & pass & +text \\
\midrule
procedural & photometric (7 lanes) & 47{,}187--47{,}190 & $\geq0.999$ & $\geq0.999$ \\
           & \texttt{width}    & 47{,}190 & 0.856 & 0.942 \\
           & \texttt{stack}    & 47{,}190 & 0.776 & 0.942 \\
           & \texttt{pressure} & 47{,}187 & 0.770 & 1.000 \\
           & \texttt{m\_pencil} & 10{,}331 & 0.430 & 0.712 \\
compose    & \texttt{compose1/2/3} & 47{,}190 & 0.964 & 0.982 \\
structural & \texttt{prefix}, \texttt{prefix\_uv} & 47{,}187 & 1.000 & 1.000 \\
           & \texttt{phase\_suffix}, \texttt{phase\_interior}
           & 47{,}086--47{,}092 & 1.000 & 1.000 \\
caption    & content (4 rungs) & 40{,}252 & $\geq0.999$ & -- \\
           & order (4 rungs)   & 40{,}252 & 1.000 & -- \\
\bottomrule
\end{tabular}
\end{table}

\clearpage
\section{Order-encoding ablation}
\label{app:codec-ablation}

Order could instead be packed into one monotone RGB channel. We bracketed eight such encoders against the HSV scheme, decoder-free: matched nearest-neighbor decode through the frozen VAE, scored by Spearman $\rho$ against true order (Table~\ref{tab:codec-ablation}). Three findings decided the design. Normalization across the slow channel's full range is the only lever: the unnormalized odometer collapses because order sits in a tiny-range channel the VAE swamps, while base-$N$ coarsening changes nothing. Fixed-arc sinusoidal codes self-intersect once total arc exceeds half the coarsest wavelength, folding order -- a representational failure, not a decoder limitation. And single-axis packing destroys within-stroke order ($\rho \approx 0.05$ through the VAE, from $1.0$ clean) because fine order rides the low-order bytes the VAE removes first; HSV ties the best single-axis encoder cross-stroke and keeps within-stroke order readable via its dedicated low-frequency $V$ channel. These frozen-VAE ceilings guided the codec choice and predate the order-native decoder, which reads the chosen code far more precisely (Appendix~\ref{app:ceiling}).

\begin{table}[h]
\centering
\caption{Coarse-order survival through the frozen VAE: matched nearest-neighbor decode, Spearman $\rho$ vs.\ true order ($n=36$ holdout; HSV rows $n=8$). Clean decode is $\approx 1.0$ for all encoders except the fixed-arc codes, which self-intersect.}
\label{tab:codec-ablation}
\small
\begin{tabular}{llr}
\toprule
encoder & order axis & $\rho$ through VAE \\
\midrule
odometer, absolute base-256        & cross-stroke & 0.259 \\
odometer, normalized base-256      & cross-stroke & 0.821 \\
normalized base-64 / 48 / 32       & cross-stroke & 0.831 / 0.833 / 0.833 \\
sinusoidal phase (cyc)             & cross-stroke & 0.842 \\
fixed-arc sinusoidal, $\lambda_{\max}{=}26$k / $65$k px & cross-stroke & 0.608 / 0.557 \\
\midrule
HSV (ours), hue                    & cross-stroke & 0.843 \\
HSV (ours), value                  & within-stroke & 0.634 \\
\bottomrule
\end{tabular}
\end{table}

\section{Raster-code diagnostics}

\begin{table}[h]
\centering
\caption{Visible-channel diagnostics after 8-bit RGB quantization. Stroke survival: fraction of source strokes retaining a visible pixel after $z$-buffering.}
\label{tab:codec}
\small
\begin{tabular}{lrr}
\toprule
& mean & tail / worst \\
\midrule
saturation-mask IoU                    & 1.000000 & 1.000000 \\
visible global-arc MAE                 & 0.000268 & 0.000824 (p99) \\
visible global-arc Spearman $\rho$     & 0.999999 & 0.999999 \\
visible within-stroke-arc MAE          & 0.001965 & -- \\
stroke survival after $z$-buffer       & 0.9928   & 0.8667 (worst) \\
\bottomrule
\end{tabular}
\end{table}

\clearpage
\section{Additional figures}
\label{app:figures}

\begin{figure}[h]
\centering
\includegraphics[width=0.78\linewidth]{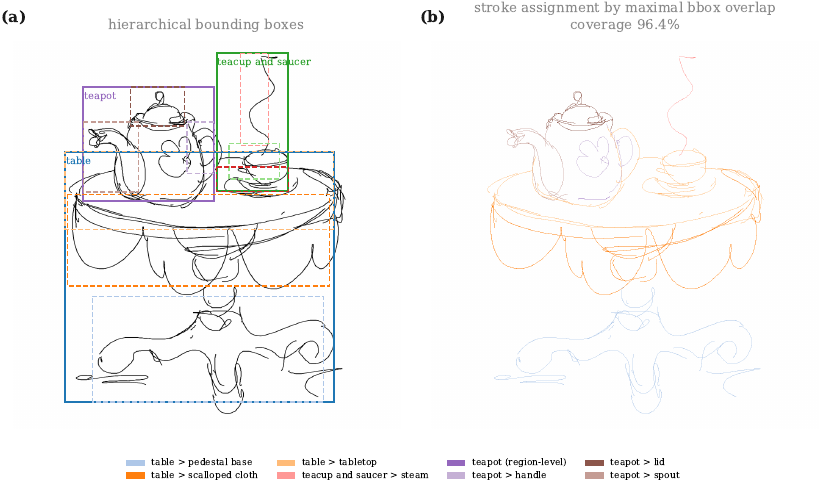}
\caption{Hierarchical annotation on one sketch. (a) Solid boxes: regions; dashed: subjects. (b) Strokes colored by assigned label.}
\label{fig:annotation}
\end{figure}

\begin{figure}[h]
\centering
\includegraphics[width=0.9\linewidth]{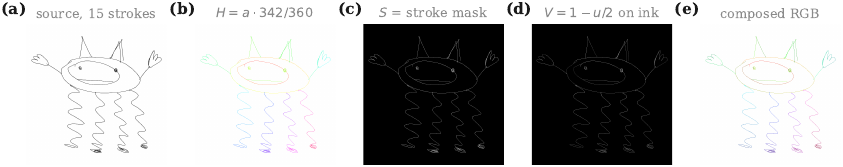}
\caption{The encoding, on one held-out drawing (15 strokes). (a) Source ink. (b) $H = a\cdot 342/360$: order as a red-to-violet sweep. (c) $S$: the visible stroke mask. (d) $V = 1-u/2$ on foreground. (e) The composed RGB target consumed by the VAE.}
\label{fig:codec}
\end{figure}

\begin{figure}[h]
\centering
\includegraphics[width=\linewidth]{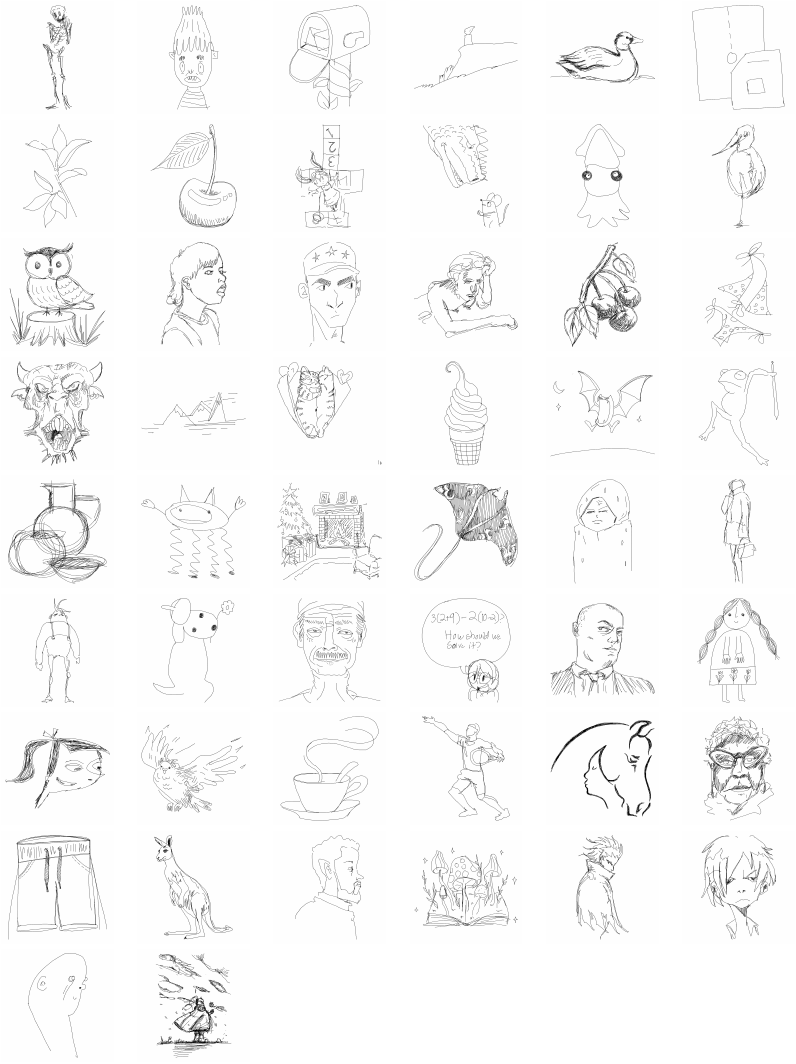}
\caption{One deterministically-sampled drawing from each of the 50 commissioned artists.}
\label{fig:artist-appendix}
\end{figure}

\end{document}